\def\year{2021}\relax
\documentclass[letterpaper]{article} 
\usepackage{aaai21}  
\usepackage{times}  
\usepackage{helvet} 
\usepackage{courier}  
\usepackage[hyphens]{url}  
\usepackage{graphicx} 
\usepackage{amsmath}
\usepackage{amssymb}
\usepackage{mathtools}
\usepackage[
bookmarks, colorlinks, breaklinks, 
	pdftitle={Alik Sokolov - vita},
	pdfauthor={Alik Sokolov},
]{hyperref}  
\hypersetup{linkcolor=black,citecolor=black,filecolor=black,urlcolor=blue} 

\usepackage{algorithm}
\usepackage{algpseudocode}
\usepackage{multirow}
\usepackage{multicol}
\usepackage{nccmath}
\usepackage{latexsym}
\usepackage{ragged2e}

\usepackage{natbib}  
\usepackage{caption} 
\title{Domain Specific Fine-tuning of Denoising Sequence-to-Sequence Models for Natural Language Summarization }

\author{Brydon Parker$^1$ \hspace{10pt} 
        Alik Sokolov$^2$ \hspace{10pt} 
        Mahtab Ahmed$^3$ $^5$ \hspace{10pt}
        Matt Kalebic$^4$ \hspace{10pt}\\
        Sedef Akinli Kocak$^5$ \hspace{10pt}
        Ofer Shai$^1$ \hspace{10pt}
                \vspace{1mm}\\
    \normalsize{$^1$ Deloitte} {$^2$ University of Toronto, Risklab} {$^3$Western University} {$^4$PwC Canada} {$^5$Vector Institute} }
        
\begin{document}

\maketitle

\begin{abstract}
Summarization of long-form text data is a problem especially pertinent in knowledge economy jobs such as medicine and finance, that require continuously remaining informed on a sophisticated and evolving body of knowledge. As such, isolating and summarizing key content automatically using Natural Language Processing (NLP) techniques holds the potential for extensive time savings in these industries. We explore applications of a state-of-the-art NLP model (BART), and explore strategies for tuning it to optimal performance using data augmentation and various fine-tuning strategies. We show that our end-to-end fine-tuning approach can result in a 5-6\% absolute ROUGE-1 improvement over an out-of-the-box pre-trained BART summarizer when tested on domain specific data, and make available our end-to-end pipeline to achieve these results on finance, medical, or other user-specified domains. Github can be found \textcolor{blue}{\href{https://github.com/VectorInstitute/Vector_NLP_Domain-Summ}{here}}, and the dataset can be found \textcolor{blue}{\href{https://www.kaggle.com/vectorinstitute/domainspecific-reddit-data-medical-and-financial}{here}}.

\end{abstract}
\section{Introduction}
As the scale and variety of text data encountered in contemporary life continues to surge, effective summaries of information have become increasingly important. However, making the choice of what information to retain represents a substantial challenge that is both costly and time-consuming when performed manually, often making automation a requirement. Valuable text data often requires domain-specific knowledge to discern, which makes it difficult to crowd-source the parsing of such texts. This need for specialized readers or labelers can make the costs of natural language processing (NLP) projects on highly technical domains prohibitive. Motivated by recent advancements in NLP and model pre-training, our work seeks to advance automation of domain-specific text summarization through elucidating optimal end-to-end methodologies, from data collection to model fine-tuning.

Automated summarization is key opportunity in the field of NLP research for a number of industries, and especially in finance and healthcare. In finance, recent industry trends in investment research have emphasized this area in particular. Historically, highly specialized and experienced investment analysts have spent much of their time synthesizing structured reports from disparate news, financial statement and investor call data. Given advances in machine learning today coupled with regulatory changes and competitive forces, industry research teams are making a rapid push for automation in this area \cite{wigglesworth2019}. 

Similarly, healthcare is an area where a significant research burden is placed on highly qualified, educated, and experienced practitioners. As an example, the recent COVID-19 crisis alone has created the need to parse through a vast amount of scholarly articles. As of April 2019, roughly one month after the novel coronavirus was declared a pandemic by the World Health Organization (WHO), the Allen Institute for AI published a dataset of over 138,000 scholarly articles referencing COVID-19, SARS-CoV-2, and related coronaviruses \cite{Wang2020CORD19TC}. When considering the need for healthcare practitioners to balance the necessity of remaining up-to-date against the sheer scale of information published, it is clear that NLP text parsing and summarization capabilities can be highly beneficial. 

While extensive progress has been achieved in methods for performing extractive and abstractive summarization \cite{allahyari2017text} \cite{summSurvey} \cite{lewis2019bart}, reliable and sufficiently large datasets for model training and evaluation have remained relatively scarce. Efforts to date have largely focused on constructing large general annotated corpora based on news, such as DUC \cite{harman-over-2004-effects}, Gigaword \cite{napoles-2012} \cite{chopra-etal-2016-abstractive}, CNN / Daily Mail \cite{nallapati2016abstractive}, and NEWSROOM \cite{grusky2018newsroom}, as well as from knowledge bases such as WikiHow \cite{koupaee2018wikihow} and longer-form documents such as research papers \cite{cohan2018discourseaware}. Recognizing the gap in specialized domain-specific datasets, our work leverages the unique community structure of Reddit, a prominent social news website that promotes discussion in forums ("subreddits") devoted to specific topics. We capitalize on the use of "TLDR" acronyms ("too long; didn't read"),  which indicate a concise summary of a lengthy comment. In this paper, we propose an improved extraction of Reddit TLDR's relative to baseline methods \cite{volske2017}. We also provide tools for end-users to develop domain-specific text summarizers. We do this by sharing our framework for generating domain-specific training data by targeting specific subreddits, thereby facilitating fine tuning of pre-trained models and enhancing their overall performance

Lastly, we provide an analysis of abstractive summarization on datasets generated from finance and medical subreddits using Bidirectional and Auto-Regressive Transformers ("BART") \cite{lewis2019bart}, a denoising autoencoder for pretraining sequence-to-sequence models that adopts and expands on Bidirectional Encoder Representations from Transformers ("BERT") \cite{devlin2018bert} and GPT pretraining \cite{radford2018improving} schemes. 

\section{Related Work}
Approaches to text summarization can be classified into two main categories: \textit{extractive} and \textit{abstractive} \cite{allahyari2017text}. Extractive summaries identify which sections of the text to retain versus discard, producing a cropped and stitched version of the original text with no alteration to phrasing. A classical approach to extractive summarization is using unsupervised methods such as Latent Semantic Analysis (LSA) \cite{ozsoy2011} whereby semantic relationships are identified through observing commonality of words across documents in a corpus. Such approaches are limited to verbatim extraction, which limits the fluidity of outputs produced. Modern NLP methods primarily focus on abstractive summarization using sequence-to-sequence deep learning models \cite{sutskever2014sequence} \cite{abstractiveSummNN2015}, which result in summaries that often contain new language not present in the original text.

Rapid progress has been made in recent years improving encoder-decoder architectures for text summarization. Convolutional or recurrent neural networks such as the sequence-to-sequence method proposed by Sutskever, Vinyals, and Le \cite{sutskever2014sequence} have greatly imporoved on the state-of-the-art at the time, while the attention mechanism-based Transformer architectures \cite{vaswani2017attention} have quickly risen to prominence for their performance and scalability. The Transformer-based GPT-2 released by OpenAI \cite{Radford2019LanguageMA} has proven the performance of unconditional language models on summarization tasks \cite{s2019extractive}, and hybrid approaches such as BART \cite{lewis2019bart} have sought to merge the bi-directional encoder approach of BERT \cite{devlin2018bert} with the left-to-right decoder approach of GPT \cite{radford2018improving}. 

While model development for text summarization has rapidly progressed in recent years, training data for summarization has largely been limited to news datasets such DUC, Gigaword, CNN / Daily Mail, and NEWSROOM, with some less popular non-news datasets such as those sourced from knowledge bases \cite{koupaee2018wikihow} and patent records \cite{sharma2019bigpatent}. While summarization is highly relevant in specialized domains such as medicine \cite{mishra2014text} and finance, it is difficult to find open, specialized and sufficiently large annotated datasets for fine-tuning of pretrained models. While previous literature has discussed the use of Reddit data for text summarization \cite{kim2018abstractive} \cite{volske2017}, the task of creating corpora specific to particular topics and communities represented by a subreddit or group of subreddits is a novel area addressed through our work. 

\section{Dataset Preparation}
Our dataset is constructed by scraping and cleaning data from the website Reddit which has a number of key features that make it an ideal source of data for text summarization:
\begin{itemize}
    \item The Reddit upvote system provides a natural filtering mechanism that allows us to remove poor quality posts, generating a high-quality dataset for training
    \item  Users often provide summaries of their posts and of others, indicating the summary with a "TLDR" label
    \item Wide variety of content with over 1 million subreddits \cite{tldrMining}, each of which corresponds to a distinct topic
\end{itemize}
In particular, the vast number of topics gives us the ability to build text summarizers for a wide variety of different domain specializations. We illustrate the benefits of domain specialization by focusing on the finance and medical domains. For finance, we curated a collection of subreddits (table X), and for the medical domain, we focus on the r/AskDocs subreddit, where patients often summarize their question to physicians in a TLDR.  

We build upon the work from\cite{tldrMining} enhancing it in a number of ways to increase the precision and recall of the summaries being extracted. 

\subsection{Dataset Collection}


We develop one end-to-end script that can be used to scrape Reddit, extract the TLDRs, and write results to a database for further filtering and cleaning. The TLDR extraction and scrapping of Reddit explained below are all present in said script and the python modules that script calls. The final steps in the process of dataset preparation are cleansing and filtering which take place in PostgreSQL scripts. 

For the purposes of this paper we focus exclusively on collecting self-posts which correspond to the original posts in Reddit, not the comments. Unlike the work from \cite{tldrMining}, we exclude comments to maximize the quality of our summaries, as self-posts tend to be longer and include the highest quality TLDRs. 

We identify self posts initially by searching for posts that contain both "TL" and "DR". In addition, we filter to the score (upvotes - downvotes) to be above 1 in order to remove lower quality posts. We also include functionality in our scraper to pull from an explicit collection of subreddits specified by the user.

\subsection{Dataset Cleaning}

We use the code from \cite{tldrMining} as our baseline for summary extraction with several key improvements to increase the quality and number of extractions:


\begin{itemize}
    \item {Rather than match TLDRs based on \texttt{tl.\{0,3\}dr} (allowing for up to 3 wild card characters between the "tl" and "dr" strings), we match based on \texttt{tl.\{0,1\}dr}, as we observe that the \texttt{tl.\{0,3\}dr} matching function produces many false positives. Most of these false positives originate from words containing "tl" near the end of the word (e.g. "abruptly") and the next word beginning with "dr".}
    \item {Instead of assuming the TLDR match occurs at the end if it is not matched at the very first token, we handle cases when the match occurs elsewhere in the body of the post. This removes the issue of creating massive extracted summaries which happen when a full post is mislabeled as a TLDR.
    
    This has the added benefit of significantly increasing the size of our training dataset, as TLDRs are often present in the beginning of the post, especially for some of the domain-specialized subreddits we focused on.}
    \item {We apply additional filtering of the extracted summaries by removing content examples where}     
    \begin{itemize}
        \item[$-$] {summary has fewer than 6 words}
        \item[$-$] {summary has equal to or fewer words than the content}
        \item[$-$] {content or summary is not in English}
        \item[$-$] {author is undefined}
        \item[$-$] {post was produced by a bot}
        \item[$-$] {post is deemed to be duplicate after removing non-alphabetical characters and converting all text to lower case.}
    \end{itemize}
\end{itemize}

\subsection{Dataset Statistics}

Prior to applying our additional filtering described above, our dataset includes 4.1 million summary and content pairs. After applying all of the filtering layers, the cleaned dataset consists of 1,687,257 training pairs. Table \ref{tab:table1} provides a summary of key domain-specific subreddits, as well as overall data volumes, in our final dataset.

\begin{table}[h!]
\begin{center}

\begin{tabular}{|p{1.5cm}|p{2cm}|p{3.60cm}|}
\hline
    \textbf{Dataset} & \textbf{Number of Summary Pairs} & \textbf{Key Subreddits} \\
 \hline
 \hline
  Full & 1,687,257 & r/relationships (26\%), r/tifu (3\%), r/AskReddit (2\%), r/leagueoflegends (2\%), r/trees (1\%), r/legaladvice (0.7\%) \\
  \hline
  Finance & 3,295 & r/wallstreetbets (33\%), r/investing (19\%) \\ 
  \hline
  Medical & 3,886 & r/AskDocs (100\%) \\
  \hline
\end{tabular}
\end{center}
\caption{Training Dataset Volumes}\label{tab:table1}
\end{table}

After filtering the posts in the general population, the average length of a summary is 35.52 words and the average length of content is 406.91 words. Some other key statistics for the full dataset and groupings of specialized subreddits can be found in Table \ref{tab:table2}.

\begin{table}[h!]
\begin{center}
\begin{tabular}{|p{1.5cm}|p{1.65cm}|p{1.65cm}|p{1.85cm}|}
\hline
     \textbf{Dataset} & \textbf{Average Words in Content} & \textbf{Average Words in Summary} & \textbf{Median Words in Summary}\\
\hline
\hline
  Full & 406.91 & 35.52 & 25 \\
\hline
  Finance & 365.80 & 33.53 & 24 \\ 
\hline
  Medical & 411.48 & 39.53 & 28 \\
\hline
\end{tabular}
\end{center}
\caption{Additional Dataset Statistics}\label{tab:table2}
\end{table}

\section{Model and Experiments}

Here we provide a summary of the models used and experiments conducted in order to determine the optimal combination of training datasets and fine-tuning methodologies, provide a summary of our results, and outline and justify the optimal fine-tuning approach. 

\subsection{Model Architecture}
We adopt the BART architecture as the initial pre-trained model, and fine-tune it using the process proposed in the original paper \cite{lewis2019bart}. The BART model is built using the standard Transformer-based sequence-to-sequence architecture \cite{vaswani2017attention}, and takes a denoising pre-training approach similar to BERT \cite{devlin2018bert}, utilizing more aggressive masking and noise functions than the original BERT pre-training. Similarly to the large version of the base model, we use 12 layers for each of the encoder and decoder, respectively, with each layer performing additional cross-attention over the final hidden state of the encoder. 

\subsection{Experiments}
Our main goal is to quantify the benefit of fine-tuning to inject additional domain knowledge into a summarizer that combines extractive and abstractive summarization techniques. To accomplish this we perform 4 primary experiments. In each set of experiments, we evaluate the benefits of pre-training on different subsets of our data and using different fine-tuning approaches, using specialized test sets for the finance and medical by us taken us subsets of training data described in Table \ref{tab:table1}. Each dataset was randomly split into train, validation, and test sets using 60/20/20 allocations into each.

\begin {enumerate}
    \item In the first experiment we apply the out-of-the-box BART summarizer that was trained explicitly on the CNN/Daily Mail datasets to our specialized test datasets directly.
    \item In the second experiment, we begin with the out-of-the-box BART summarizer, and use the full Reddit dataset for fine-tuning. In this experiment, we only train the decoder, while keeping all the encoder layers frozen.
    \item   In the third experiment, we follow everything that was done in the second but additionally train the encoder layers.
    \item  In our final set of experiments, we take the out-of-the-box model and fine-tune it directly on each domain specific training dataset (Table 1), and fine-tune both the encoder and the decoder.
\end{enumerate}

\subsection{Fine-Tuning Approach}

As is done in BART and initially in RoBERTa \cite{roBERTa} we aim for very large batch sizes and accomplish this using gradient accumulation. We hypothesize that large batch sizes are of particular importance for Reddit data due to the noise in this dataset. Managing noise is a challenge common to training summarization models, which often causes peak performance to be achieved very early in the training process. In most of our experiments, we also see imporvement only for the first few epochs of fine-tuning. This is consistent whether fine-tuning on the entire Reddit dataset, or one of our specialized subsets. We expect this is due to the better quality summaries that are present in the CNN/Daily Mail dataset, and the high quality of the out-of-the-box language model. During training we set the maximum number of attempts for validation loss to increase equal to 1 as our stopping criterion. We do this to conserve computational resources, as due to our very large batch size the validation loss steadily decreases after it initially starts going down.

For autoregressive generation parameters we match the original BART paper where appropriate, matching their length penalty, number of beams, minimum number of output tokens, and removal of duplicated tri-grams. We set the maximum number of output tokens to 120 (resulting in average output length of approx. 80 words for all of our models), maximum number of input tokens at 512, and use nucleus sampling \cite{Holtzman2020TheCC} with Nucleus p = 0.95 as is proposed in the paper. For additional details of hyperparameters and their tuning refer to Table \ref{tab:table3}.

\medskip

\begin{table}[h!]
\begin{center}
\begin{tabular}{|p{4cm} | p{3.1cm}|}
\hline
    \textbf{Hyperparameter} & \textbf{Indicative Range Tested} \\
 \hline
 \hline
  Learning Rate & $\mathbf{1e^{- 4}}$ / $1e^{-5}$ / $1e^{- 6}$ \\
  L2-Lambda  & $\mathbf{0.001}$/ $0.01$ / $0.1$ / $1.0$ \\
  Length Penalty & $\mathbf{1.0}$/ $2.0$ / $10.0$\\
  Batch Size & $512$/ $\mathbf{1024}$ / $8000$\\
  Number of Beams & $5$ \\
  Repetition Penalty & $2.5$ \\
  Max Output Tokens & $120$ \\
  Min Output Tokens & $56$ \\
  Max Input Tokens & $128$/ $256$ / $\mathbf{512}$ \\
  Nucleus P & $0.95$\\
  Learning Rate Scaling & $0.95$ \\ 
\hline
\end{tabular}
\end{center}
\caption{Hyper-parameters used for the experiments (in boldface) and the ranges that were searched during tuning.}\label{tab:table3}
\end{table}

\normalsize

A cluster with one NVIDIA Titan XP GPU, 8 vCPUs and 12 GB of host memory is used to train the models. We expect that by scaling GPU memory and increasing model size, e.g. through raising the maximum sequence length to account for longer-term dependencies in the input.

\medskip

\section{Results}

We find that the highest ROUGE scores on our test dataset are obtained using the fine-tuning approach in our fourth set of experiments, where we fine-tune on the specialized datasets directly.

We find that for training the model on the full corpus that freezing the encoder led to a significant improvement in the model performance on each domain, resulting in a much better performance on our test set in the second set of experiments compared to the third set of experiments (see Table \ref{tab:table4}). For the domain-specific summarizers, no layer freezing led to the best outcomes. We believe this indicates that allowing for encoder adaptation to the vocabulary can lead to additional improvements when training on a new specialized domain.

\subsection{Model Evaluation}

We use ROUGE-1 recall, ROUGE-2 recall, ROUGE-1 precision and ROUGE-2 precision as our primary evaluation metrics. In order to generate the summaries, we use teacher forcing for training and autoregressive generation for evaluation. We generally match the BART architecture with most of our experiments focused on optimizing fine-tuning. We experimented with freezing different sections when training both for the encoder and the decoder to control learning rates across layers, and find that different training tasks performed best with different layer freezing setup.
 
Table \ref{tab:table4} summarizes our experimental results. We hypothesize that the improved performance from encoder freezing for the full corpus model and not the domain specific models is due to the quality and type of the summaries that are most common in the full corpus. In the full corpus most of our summaries are from subreddits that use very different language than our domain-specific ones, potentially making the encoder fine-tuning counterproductive; . The models fine-tuned on all of Reddit data without any layer freezing (experiment 3) did not perform significantly better than the out-of-the-box pre-trained model on specialized domains. Significant gains on the domain specific datasets are achieved from training on the general dataset with the encoder frozen, and even further gains from fine-tuning on each specific domain.


\begin{table*}[h!]
\begin{center}
\begin{tabular}{|p{8.79cm}|p{1.5cm}|p{1.5cm}|p{1.5cm}|p{1.5cm}|}
\hline
    \textbf{Model} & \textbf{R1} & \textbf{R2} & \textbf{P1} & \textbf{P2} \\
 \hline
 \hline
  1. Train - CNN/DM & $31.37$ & $3.62$ & $11.18$ & $1.19$\\ 
  \hline
  2. Train - CNN/DM; Fine-tune - General; Encoder Frozen & $35.70$ & \pmb{$4.79$} & \pmb{$12.39$} & \pmb{$1.57$}\\
  \hline
  3. Train - CNN/DM; Fine-tune - General; Fully Unfrozen& $31.69$ & $3.76$ & $11.93$ & $1.34$\\
  \hline
  4. Train - CNN/DM; Fine-tune - Finance; Fully Unfrozen& \pmb{$36.94$} & $4.54$ & $12.18$ & $1.42$\\ 
  \hline
\end{tabular}
\end{center}
\caption{Experiment Results}\label{tab:table4}
\end{table*}

We find reddit data alone to be insufficient to train a full summarization model. This is clear when considering how heterogeneous language, as well as format and quality of summaries, are across subreddits, making it a less attractive dataset for training a base summarizer than traditional datasets like CNN / Daily Mail. This makes fine-tuning a pre-trained summarizer using our domain specialized datasets the better strategy compared to training a specialized summarizer from scratch.

As the corresponding ROUGE scores in Table \ref{tab:table4} indicate, summarization quality improves greatly as more specialized data is used for training summarizers. This can also be seen qualitatively, with the more specialized models learning to tailor their summaries more to the specific domain. As we discuss in the qualitative observations section below, the structure over the Reddit TLDR summaries also offers some advantages over the CNN / Daily Mail dataset, which in many cases leads to potentially more useful summaries.

\subsection{Qualitative Observations}

We give some examples of summaries generated by the best models from experiments 1,3, and 4 (omitting the results of experiment 2 where models fine-tuned on the full reddit data with both the encoder and decoder trained). We also omit the original text of summaries for brevity and to protect Reddit users' privacy, but discuss the quality of each summary separately.

\medskip

\begin{table*}[h!]
\begin{center}
\begin{tabular}{|p{4.7cm}|p{5cm}|p{5cm}|}
\hline
 \textbf{Out-of-the-Box Model} & \textbf{General Reddit Fine-Tuning} & \textbf{Specialized Fine-Tuning}\\
 \hline
Had some chest pain last night but things don't add up. No longer experiencing this issue, curious what it might have been ¬†and how to find out more about Ehlers-Danlos and other conditions that may cause the same symptoms in a person of similar age or different health issues than those I am currently suffering from! Have you had any unusual experiences with your body? If so please share them on our Facebook page at: \url{http://soulmatestories@dailymailonline} &  I'm a 24 year old Caucasian female BMI of 20115 pounds and 5'3.5 I have asthma, Ehlers-Danlos or some other condition that causes chest pain?  If it wasn't an issue in the past can you tell me what might be going on with my heart attack last night - It's not like there is any clotting risk factors aside from Nuvaring for birth control so why did this happen after dancing around all day long to no one else who &  had pain in my chest and I am sure this couldn't have been a heart attack,  what if it was something else that would make me feel better? What do you think is going on with the tachycardia around 115 BPM or should be an issue for someone who has no clotting risk factors aside from Nuvaring to worry about after dancing last night there's not enough blood pressure readings so maybe its just low sodium levels because of being at work every 30 minutes - can\\
 \hline
 Liver MRI-Focal Nodular Hyperplasia or something else? Hello docs of reddit! I'm hoping you guys can help me out. A couple months ago i went to  the doctor for my psoriatic arthritis diagnosis and a ct scan showed two liver lesions that are nonspecific, but definitely not hemangiomas¬†probably F focal nd narcolepsy/hyperparathyroidism (HPC) The doctors suggested follow up imaging with either & Liver MRI-Focal Nodular Hyperplasia or something else? I'm not sure if some prior info would be helpful to you guys. A couple months ago  got my psoriatic arthritis diagnosis and am still in a lot of pain, but don't want the biopsy just because it's possible that its Hepary Adenoma/Hemangiomas - can someone help me out with this one please! If anyone has any information about liver imaging for people who & I  I'm not sure if some prior info would be helpful to you guys? liver MRI-Focal Nodular Hyperplasia or something else that my doctor can do follow up imaging and a biopsy on the 2.1 cm lesion along with an unmeasured but smaller lesions are nonspecific, maybe it's Hepatic Adenoma for another child soon - don't want all these tests just because of what they think is going in there probably no reason why this should\\ 
 \hline
\end{tabular}
\end{center}
\caption{Sample Results - Health Summarizer}\label{tab:table5}
\end{table*}

\normalsize

In both of the examples in Table \ref{tab:table5} the specialized fine-tuning model picks out relevant medical facts from the original summary. Interestingly, the specialized summarizer also picks up on the fact that the chest pain was specifically not a heart attack, which both of the other summarizers fail to do. We see a similar pattern from the second example in Table \ref{tab:table5}, where the medical specific summarizer extracts the fact that in the follow up session the doctor can do additional imaging or a biopsy on the lesion, which the first two summarizers also fail to do.  

Table \ref{tab:table6} we see some examples from where the specialized summarizer creates summaries similar in language to that used in r/wallstreetbets, one of the most prevalent finance subreddits in our corpus. This serves as an illustration for some caveats of using Reddit data for NLP domain adaptation.

\begin{table*}[h!]
\begin{center}
\begin{tabular}{|p{4.7cm}|p{5cm}|p{5cm}|}
\hline
 \textbf{Out-of-the-Box Model} & \textbf{General Reddit Fine-Tuning} & \textbf{Specialized Fine-Tuning}\\
 \hline
 \hline
The efficient market hypothesis supposes that an individual can't beat the markets. Most advice floating around rfinance andrinvesting is with regards to long-term investing through a range of ETFs, writes John Defteriosa in his column for The Motley Fool's "Moneyquiz" series on Sunday nights at 10pm ET/PT (11:00 pm BST) If most people believe in this strategy then why haven't more shrewd investors figured out how &   Question  ETFs and efficient markets. Why haven't the more shrewd investors figured out a way to essentially capture this future growth via arbitrage? If most people believe in that strategy, why isn't all of it priced-in into these investment strategies already at risk-reward trade off? The idea is an investor enjoys full diversification and low fees on their investments through exchange traded funds (ETF's) What do you think about whether or not there are enough opportunities &   why aren't all of the growth priced in to these ETFs already?  Why is this a good strategy, and isn't arbitrage better than an individual investing through them on long-term risk-reward trade off. If I can buy my way into it now would be great for me but not you or your investor if they think that's what everyone else does as well - by going with short term returns over 50\% market return + low fees will go both ways!\\
 \hline
\end{tabular}
\end{center}
\caption{Sample Results - Finance Summarizer}\label{tab:table6}
\end{table*}

\normalsize
Although both the General Reddit and Specialized model are able to capture a lot of the financial context of the underlying post, the finance example also showcases some of the pitfalls of domain specialization. Given the prevalence of r/wallstreetbets data in our test set, the specialized model achieves higher ROUGE scores by biasing towards much of the colloquial langauge present on that subreddit, and less so on finance specific text. Nevertheless, both of these models perform much better qualitatively on the financial context of text data than the out-of-the-box CNN / Daily Mail model.

Overall, we see a number of qualitative advantages for our fine-tuned models across both of our specialized domains. One common theme from the out-of-the-box model is that it often ends in recommending a URL, sending an email to someone, or sharing a perspectives on social media. Qualitatively, the out-of-the-box summarizer also focuses a lot more on the narrative, generating summaries that are well-written and appear intriguing, but omit key facts, which can also be seen in Table \ref{tab:table5} and Table \ref{tab:table6}. We believe the Reddit summarizers extract these additional details more effectively as the training summaries from Reddit are better designed to do so, as they are intended to serve as a replacement for reading the post, rather than a "hook" to get the reader to engage with the content.

\section{Significance and Future Work}

Our research demonstrates the potential application of our data extraction and fine-tuning pipeline in the fields of finance and healthcare. Models fine-tuned in a manner described in our paper can be used as standalone text summarization models, or further improved by augmenting training data with additional public or proprietary data sources. More importantly, we show a clear positive trend in the quality of text summarization for a large language model such as BART when fine-tuned using specialized data, even in cases where the amount of labeled examples available for fine-tuning is relatively small. We also quantify the relative extent to which additional specialization can improve the quality of summarization. We hope these data points can be used by industry practitioners to help with the cost-benefit analysis for selecting the right level of fine-tuning, from using large out-of-the-box training models to obtain potentially costly high-quality specialized data.   

Based on the qualitative evidence of domain specialization and the increase in the ROUGE-1 scores from specialization over the general model with the encoder frozen we conclude that the domain specific models are able to specialize into their respective domains. The models do this by learning what parts of the text are more significant based on the domain, and learning to more effectively leverage relevant vocabulary when developing abstractive summaries. Due to these observations we believe domain adaptation capabilities are an excellent use of the dataset we have created. Those who use this dataset for the purpose of domain-specialization should make sure the language and discourse in these subreddits aligns well with the business problem at hand.

In addition to the domain-specialization use case, due to the increase in the performance and summary quality on each domain from training on the entire Reddit dataset with encoder frozen, we also believe there is value in using this dataset to improve general summarization models. Those interested in using this dataset for general summarization should note the importance of layer freezing to avoid the model picking up unwanted language and patterns from general Reddit data.

The summaries dataset can be extended in a number of ways to provide additional value. There is a large number of TLDRs that summarize website content linked to in Reddit posts, and scraping the content of these website can augment the summaries dataset with high-quality data. Summaries can also be augmented by parsing through comment trees and extracting TLDRs from long comments. 

The summarization models can also be improved in several ways. A high-potential area of improvement is controlling catastrophic forgetting, via regularization schemes such as layer freezing or varying learning rates across layers. Forgetting can also be addressed through modifying training data, by augmenting fine-tuning data with samples of the original summaries. Finally, new architectures like XLNet \cite{Yang2019XLNetGA} can be employed to increase the max sequence length the transformer encoder-decoder can handle to improve performance on very long documents. 

\section*{Acknowledgements}
We want to thank Vector Institute industry sponsors, researchers and technical staff who participated in the Vector Institute's NLP Project (\url{https://vectorinstitute.ai/wp-content/uploads/2020/12/nlp-report-final.pdf}).

\bibliography{ref}
\end{document}